# Generating grid maps via the snake model

Zhiwei Wei, Nai Yang, Wenjia Xu, Su Ding

**Abstract**—The grid map, often referred to as the tile map, stands as a vital tool in geospatial visualization, possessing unique attributes that differentiate it from more commonly known techniques such as choropleths and cartograms. It transforms geographic regions into grids, which requires the displacement of both region centroids and boundary nodes to establish a coherent grid arrangement. However, existing approaches typically displace region centroids and boundary nodes separately, potentially resulting in self-intersected boundaries and compromised relative orientation relations between regions. In this paper, we introduce a novel approach that leverages the Snake displacement algorithm from cartographic generalization to concurrently displace region centroids and boundary nodes. The revised Constrained Delaunay triangulation (CDT) is employed to represent the relations between regions and serves as a structural foundation for the Snake algorithm. Forces for displacing the region centroids into a grid-like pattern are then computed. These forces are iteratively applied within the Snake model until a satisfactory new boundary is achieved. Subsequently, the grid map is created by aligning the grids with the newly generated boundary, utilizing a one-to-one match algorithm to assign each region to a specific grid. Experimental results demonstrate that the proposed approach excels in maintaining the relative orientation and global shape of regions, albeit with a potential increase in local location deviations. We also present two strategies aligned with existing approaches to generate diverse grid maps for user preferences. Further details and resources are available on our project website: https://github.com/TrentonWei/DorlingMap.git.

## 1 Introduction

Grid maps, also recognized as tile maps, provide an effective mechanism for visualizing geographic data (Eppstein et al., 2015). These maps represent regions as uniform grids, as depicted in Figure 1, allowing for the preservation of local relationships while maintaining an overall similarity to the original map. This consistency is crucial, as it minimizes the issue of large regions dominating the map and facilitates the identification of smaller ones. Additionally, the uniform size and shape of the grids support successful data comparison and exploration, making it a suitable technique for time-series data visualization (Calvo et al., 2023). These distinct qualities distinguish them from better-known techniques such as choropleths (Egbert and, Slocum, 1992) and cartograms (Nusrat and Kobourov, 2016).

These unique qualities distinguish grid maps from better-known techniques such as choropleths (Egbert and Slocum, 1992) and cartograms (Nusrat and Kobourov, 2016). Consequently, grid maps have attracted attention from mapping professionals, enthusiasts, and popular media and some approaches are proposed recently to automate their production (Slingsby, 2018). These approaches typically involve two steps: generating a grid arrangement that closely resembles the original map and then using a one-to-one matching algorithm to assign each region to a grid. Eppstein et al. (2015) conducted a detailed analysis of the matching algorithm by considering optimization criteria of preserving location, adjacency, and relative orientation. This approach laid the foundation of a matching algorithm for subsequent approaches. However, their grid arrangement was obtained by adapting grids to the boundary's bounding box, and handling empty cells was required. Meulemans et al. (2016) then addressed empty cells using a simulated annealing algorithm but limited the applicability to regions with close-to-uniform spatial distribution. Recent approaches have focused on achieving a uniform spatial distribution to prevent suboptimal arrangements resulting from non-uniform distributions. To achieve this, McNeill and Hale (2017) displaced region centroids to equidistant positions, creating a more grid-like conf Motivated by the above thoughts, we employ the Snake model to generate a new boundary for grid fitting, thereby refining the resulting grid arrangement in grid maps. To structure the region centroids and boundary nodes in this new approach, we employ a revised constrained Delaunay triangulation (CDT) as a linear network. The forces acting on this linear network are computed to displace the region centroids as if they were not arranged in a grid-like pattern. Throughout the Snake model process, these forces are iteratively applied until an acceptable new boundary is obtained, resulting in the simultaneous displacement of both the region centroids and boundary nodes. Finally, to create the grid map, the grids are fitted to the new boundary using the one-to-one match algorithm to assign each region to a grid. This process can better maintain the relative orientation and salient local feature by comparing to McNeill and Hale's (2017) approach, thus, result in a better grid map layout.

iguration, and then transformed the boundary for grid fitting based on the displacement of these centroids. Meulemans et al. (2020) proposed to decompose the initial global shape into parts based on salient local features and area density. Grids were then generated for each sub-area using a mosaic cartogram technique that ensures uniform spatial distribution. However, despite their utility, these approaches still have some limitations. McNeill and Hale's (2017) grid map, for instance, may obscure salient local features, and the consistent maintenance of relative orientation between regions is not guaranteed. Similarly, Meulemans et al.'s (2020) method may introduce distortions in the connecting parts of sub-areas. Consequently, there remains a pressing need to explore novel approaches aimed at enhancing the production of grid arrangements for grid map visualization.

McNeill and Hale's (2017) approach is user-friendly and concise, thus, we also build our approach upon their approach. The crux of their method involves transforming the boundary based on the displacement of region centroids, requiring these centroids to be shifted in a grid-like manner, followed by fitting grids to the updated boundary. However, the displacement of centroids and boundary nodes are conducted separately in their approach which may arise from the erosion of salient local features and relative relations in the newly generated boundary. Displacement is a fundamental operation in cartographic generalization, and many algorithms have been developed to address this task. Among them, the Snake model is particularly powerful for linear network displacement because it enables global optimization of map quality criteria through energy minimization (Bader, 2001; Liu et al., 2014). If we structure the boundary nodes and region centroids as a linear network, the Snake model used in cartographic generalization can also be applied to displace the boundary nodes and region centroids simultaneously to create a new boundary. Then fitting the grids to the new boundary can help obtain a better grid arrangement to the greatest possible extent.

- *Zhiwei Wei is with Aerospace Information Research Institute, Chinese Academy of Sciences. E-mail: 2011301130108@whu.edu.cn.*
- *Nai Yang is with School of Geography and Information Engineering, China University of Geosciences(Wuhan). E-mail: yangnai@cug.edu.cn.*
- *Wenjia Xu is with Beijing Univeristy of Posts and Telecommunications. E-mail: xuwen-jia16@mails.ucas.ac.cn.*
- *Su Ding is with the College of Environmental and Resource Science, Zhejiang A & F University. E-mail: suding@zafu.edu.cn.*

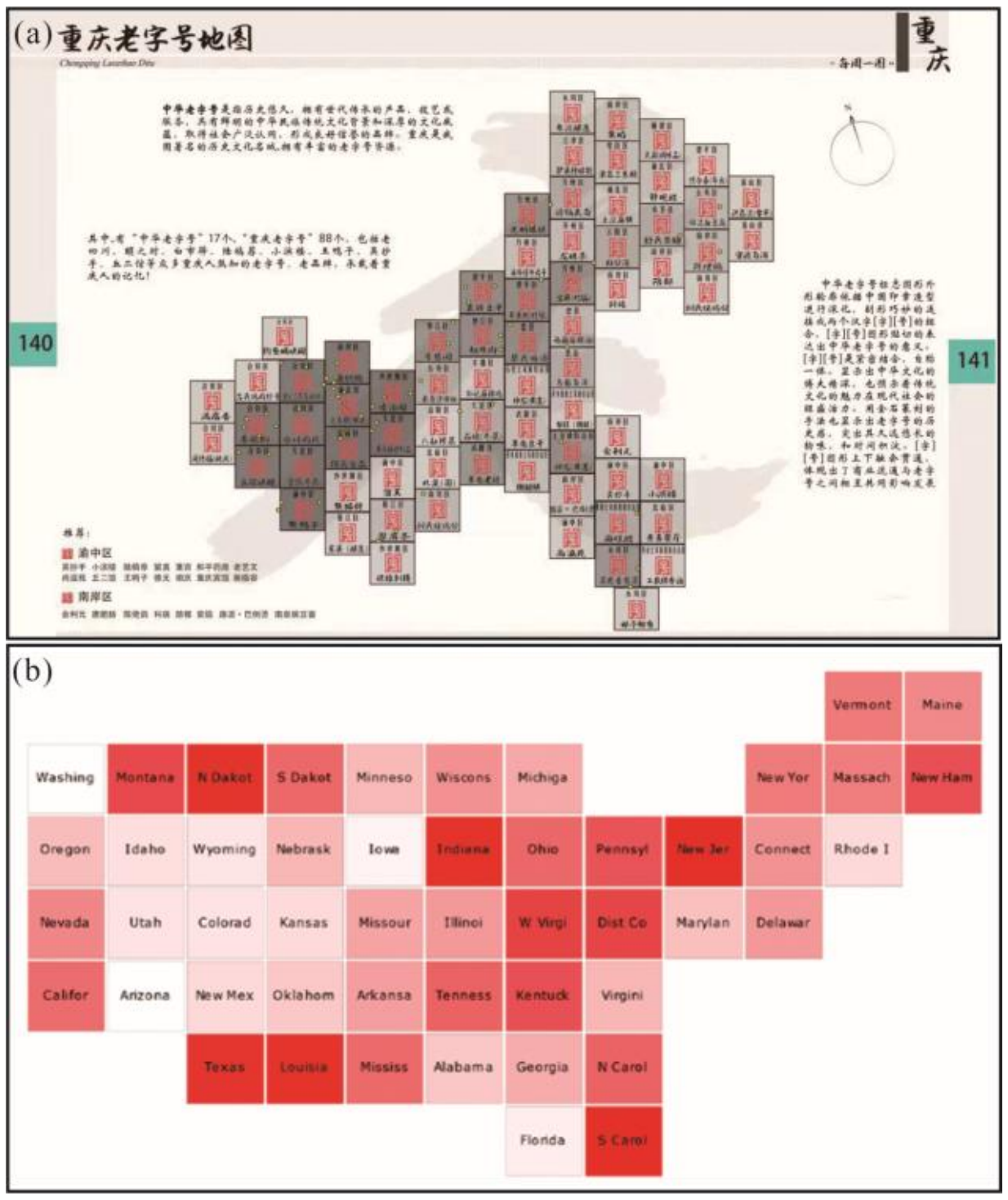

Figure 1. Examples of grid maps. (a) The grid map of old and famous brands in Chongqing, China (Su et al., 2023); (b) The grid map of the USA (McNeill and Hale, 2017).

## 2 Related works

### 2.1 The grid maps

The visualization of geographical statistics using reshaped regions, such as the Dorling map (Dorling, 1995; Inoue, 2011) or Demers cartogram (Nickel et al., 2022), has a long history, which can be dated back to The Rand McNally World Atlas of 1897 (Reyes Nunez, 2014). However, these maps use proportional shapes to represent statistical values, which can make it difficult for users to locate smaller regions. As such, grid maps with grids of the same size and shape are sometimes preferred when small regions are also important. The uniformity of these grids allows for successful comparison and exploration of data, making it suitable for time-series visualization or proximity-aware visualization (Meulemans et al., 2016; Zhou et al., 2024). With the rapid development of information technology, spatial-temporal statistics have become increasingly available in recent decades (Wei et al., 2023). This, coupled with the pressing need for time-series visualization, has propelled the popularity of grid maps. They are now widely used in popular news outlets, social media platforms, and academic literature (Slingsby, 2018; Wang et al., 2023).

The earlier grid maps were mostly produced by hand. Eppstein et al. (2015) may be the first to develop a software program to automate this process. They tackled the problem of identifying a one-to-one relationship between regions and a provided grid arrangement by considering optimization objectives that prioritize preserving location, adjacency, and relative orientation. The given grid arrangement is acquired by adapting grids to the boundary's bounding box. If the number of grids exceeds the number of regions after adaptation, a strategy for handling empty cells is required to finalize the grid arrangement. In this regard, Meulemans et al. (2016) conducted a thorough analysis of handling empty grid cells and developed a heuristic algorithm for this task. However, the grid arrangements produced by both Eppstein et al. (2015) and Meulemans et al. (2016) are only appropriate for close-to-uniform spatial distributions. To address this limitation, several improved approaches have emerged, which first generate better grid arrangements and then assign regions using Eppstein et al.'s (2015) one-to-one match algorithm. For example, McNeill and Hale (2017) displaced region centroids to ensure equidistance between nearby centroids, generating a more grid-like configuration. These displaced centroids are then used to transform the boundary in preparation for grid adaptation. However, this method creates a risk of salient local features eroding and relative orientation relation distorting in the newly generated boundary as centroid displacement and boundary node relocation are performed independently. Meulemans et al. (2020) presented a novel approach to decomposing the initial global shape into parts based on the salient local features of the boundary and area density. Grids are then generated for each sub-area using a mosaic cartogram technique which guarantees a uniform distribution of space. However, this technique may result in distortions in the connecting parts of sub-areas. Therefore, for certain new applications, the approach introduced by Eppstein et al. (2015) may still be the most suitable option. Building upon this approach, Lin et al. (2019) then proposed a puzzle grid map for temporal data visualization, while Ai et al. (2016) utilized it for population visualization.

In summary, although Eppstein et al.'s (2015) approach is commonly utilized, it is primarily suitable for nearly uniform spatial distributions. For generating grid maps for intricate geographic polygons, McNeill and Hale's (2017) method offers user-friendly and succinct characteristics. Nonetheless, their approach may overlook preserving the relative orientation relationship between grids. Therefore, there is still a need to develop a new approach capable of effectively balancing all the quality requirements of grid maps.

### 2.2 The Snake model

The Snake model was originally proposed by Kass et al. (1988) and applied in the field of computer vision. After being introduced by Burghardt and Meier (1997) into the field of cartographic generalization, it soon becomes widely used for linear objects displacement due to its advantages of small displacement propagation and good maintenance of geometry characteristics (Guilbert et al., 2016; Wang et al., 2017; Guo et al., 2017). The linear object displacements, namely the deformation of the curves, due to external forces on their nodes are described as energy. Then the optimal shape of the curve after displacement can be calculated via Finite Element Modelling (FEM) to solve the energy minimization problem, For more details about the mathematical derivation process of solving the energy minimization problem see Liu et al. (2015). The resulting displacement ($d$) of the curve can be finally represented as a matrix function, as defined in Eq. (1).

$$d = K^{-1} f \tag{1}$$

where $K$ is the element stiffness matrix of the curve, and $f$ is the force matrix of the curve.

According to FEM, $K$, $f$, and $d$ in a part (namely a segment, denote as $s$) of the curve ($K^s$, $f^s$, and $d^s$) in the direction of $x$ can be computed as Eqs. (2) and (3), (and the direction of $y$ can be calculated similarly).

$$K^s = \begin{bmatrix} \frac{6\alpha h^2 + 60\beta}{5h^3} & \frac{\alpha h^2 + 60\beta}{10h^2} & \frac{-6\alpha h^2 - 60\beta}{5h^3} & \frac{\alpha h^2 + 60\beta}{10h^2} \\ \frac{\alpha h^2 - 60\beta}{10h^2} & \frac{2\alpha h^2 + 60\beta}{15h} & \frac{-\alpha h^2 - 60\beta}{10h^2} & \frac{-\alpha h^2 + 60\beta}{30h} \\ \frac{-6\alpha h^2 - 60\beta}{5h^3} & \frac{-\alpha h^2 - 60\beta}{10h^2} & \frac{6\alpha h^2 + 60\beta}{5h^3} & \frac{-\alpha h^2 - 60\beta}{10h^2} \\ \frac{\alpha h^2 + 60\beta}{10h^2} & \frac{-\alpha h^2 + 60\beta}{30h} & \frac{-\alpha h^2 - 60\beta}{5h^3} & \frac{2\alpha h^2 + 60\beta}{15h} \end{bmatrix} \tag{2}$$

$$d^s = \begin{bmatrix} d(x_0) \\ d'(x_0) \\ d(x_1) \\ d'(x_0) \end{bmatrix} \qquad f^s = \begin{bmatrix} \frac{1}{2} hf(x_0) \\ \frac{1}{12} h^2 f(x_0) \\ \frac{1}{2} hf(x_1) \\ -\frac{1}{12} h^2 f(x_1) \end{bmatrix} \tag{3}$$

where $h$ is the length of $s$, $x_0$, and $x_1$ are the coordinates of two endpoints of $s$ in the direction of $x$; accordingly, $d(x_0)$ and $d(x_1)$ are the displacement magnitude of two endpoints in the direction of $x$; while $d'(x_0)$ and $d'(x_1)$ are the first-order derivative of two endpoints in the direction of $x$; $f(x_0)$ and $f(x_1)$ represent the component of force on two endpoints in the direction of $x$. $\alpha$ and $\beta$ are the parameters that describe the material properties which decide the elasticity and stiffness of snake model and can be set by users.

## 3 Quality criteria

We aim to produce grid maps without gaps in this approach. As illustrated in previous works (Eppstein et al., 2015; McNeill and Hale, 2017; Meulemans et al., 2020), some properties (e.g., location, shape, and topology) of the original data need to be preserved as much as possible and are summarized as follows.

**(1) Geometry**

**Local location maintenance**: the distance between the region centroids and the grid centroids should be as small as possible, (McNeill and Hale, 2017).

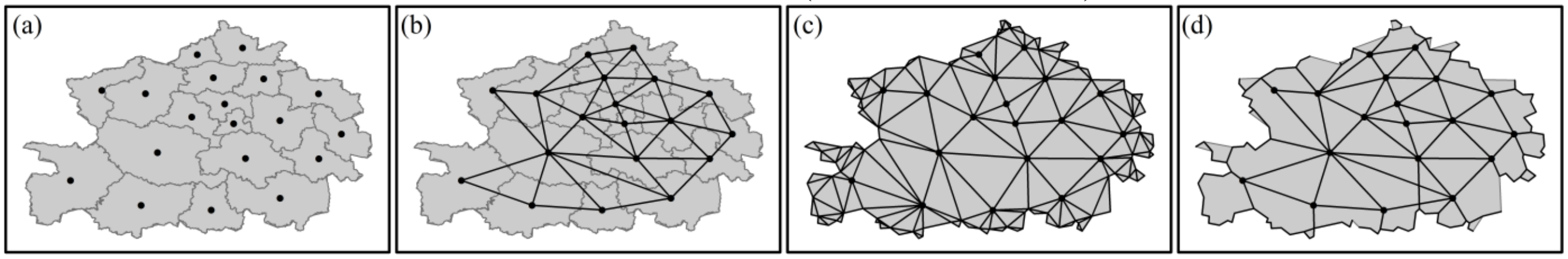

Figure 2. Construction of the linear network. (a) The regions; (b) The adjacent relations between regions; (c) The CDT for region centroids and boundary nodes; (d) The revised CDT.

**Global shape maintenance**: the overall appearance of the grid arrangement should resemble its containing shape (McNeill and Hale, 2017).

**Global size maintenance**: the total area of all grids should be close to that of all regions (McNeill and Hale, 2017).

**(2) Relation**

**Topology relation maintenance**: regions that are adjacent to each other should be assigned to adjacent grids and vice versa (Eppstein et al., 2015).

**Relative relation maintenance**: the relative relations between regions should be similar to those between the corresponding grids, specifically with respect to the relative directions (Meulemans et al., 2020).

Two things need to be mentioned. First, some other properties may also be considered in practice but not noted here, e.g., global location (McNeill and Hale, 2017) or contiguity (Meulemans et al., 2016). The reason is that global location can be explicitly maintained with relative relation maintenance, and contiguity can be expressed as specially defined topology relations. Second, the properties of the original data will inevitably be changed due to a new representation of identical grids. Thus, the quality requirements in grid maps are not like those in the database domain which should be satisfied completely, they only need to be optimized or fulfilled as much as possible.

## 4 Methodology

### 4.1 Framework

In this study, the Snake model is employed to concurrently displace region centroids and boundary nodes, generating a new boundary for grid fitting. The utilization of the Snake model necessitates the establishment of a structured linear network and the computation of forces. Therefore, our approach consists of four steps, as follows.

**Step 1**. Construct a linear network structuring the region centroids and boundary nodes;

**Step 2**. Compute forces acting on the established linear network;

**Step 3**. Iteratively apply the Snake model to the linear network, assigning forces to displace region centroids and boundary nodes until a satisfactory new boundary is achieved.

**Step 4**. Fit grids to the newly generated boundary and assign regions to grids.

The details for these steps are illustrated in the subsequent sections.

### 4.2 Step 1. Linear network construction

Suppose the regions are $\{Reg_m\}_{m=1}^{M}$, their centroids are $\{rv_m\}_{m=1}^{M}$, $rv_m$ is the centroid of region $Reg_m$, the boundary polygon of $\{Reg_m\}_{m=1}^{M}$ is $P_{\mathrm{regs}}$, the nodes of $P_{\mathrm{regs}}$ are $\{bv_n\}_{n=1}^{N}$. To displace $\{rv_m\}_{m=1}^{M}$ and $\{bv_n\}_{n=1}^{N}$ simultaneously using the Snake model, a linear network structuring these elements is constructed in accordance with the quality requirements outlined in Section 3.

First, the constraint of relative relation maintenance requires that the relative relationships between regions should be preserved, it is imperative that the linear network accurately reflects the inherent relations between these regions. The relative neighbor graph (RNG), a subgraph derived from the constrained Delaunay triangulation (CDT), is an effective means of expressing the spatial distribution of objects (Wei et al., 2018). Following the construction of a CDT, we can use the RNG to establish the linear networks for $\{rv_m\}_{m=1}^{M}$ and $\{bv_n\}_{n=1}^{N}$ by eliminating unnecessary edges. Second, it is necessary to ensure that the grid arrangement coincides with the enclosing shape to uphold the global shape maintenance constraint. Therefore, during the linear network construction, we consider the boundary polygon $P_{\mathrm{regs}}$ as a constrained polygon in CDT construction. As $\{bv_n\}_{n=1}^{N}$ signifies the boundary nodes, it is clear that we integrate the boundary constraints in CDT construction, as displayed in Figure 2(c). Thirdly, in all linear network constructions, it is crucial to consider and preserve topology relation maintenance constraints. Hence, when we remove edges from the CDT and obtain the RNG, we refrain from deleting edges that connect adjacent regions.

Consequently, the linear network is constructed as follows. Initially, a CDT is first built for $\{bv_n\}_{n=1}^{N}$ and $\{rv_m\}_{m=1}^{M}$ in which $P_{\mathrm{regs}}$ is taken as a constrained polygon, as shown in Figure 2(c). Subsequently, edges are removed from the CDT to obtain the RNG; Notably, the edges connecting adjacent regions, as depicted in Figure 2(b), are preserved during this process. The resulting RNG is shown in Figure 2(d).

### 4.3 Step 2. Force computation

The forces on the linear network should promise that the region centroids $\{rv_m\}_{m=1}^{M}$ are equidistant (more grid-like) after the displacement. Because the relative orientation of neighbors should be preserved for $\{rv_m\}_{m=1}^{M}$, the desired location $rv'_m$ for a given region centroid $rv_m$ can be approximately computed according to McNeil and Hale (2017) as Eq. (4).

$$rv_m^{new} = \frac{1}{|N(i)|} \times \sum_{j \in N(i)} (rv_m^{old} + S\hat{u}_{ij}) \tag{4}$$

where $N(i)$ is the region centroid set of neighbors of $rv_i$, the neighbors are determined by the built linear network in Section 4.2. $S$ is the estimated grid size and is defined as Eq. (5) according to the global size maintenance.

$$S = \sqrt{\frac{A}{M}} \tag{5}$$

where $A$ is the area of all regions, and $M$ is the grid count. $\hat{u}_{ij}$ is the unit displacement vector between two centroids $rv_i$ and $rv_j$, and is defined as Eq. (6).

$$\hat{u}_{ij} = \frac{rv_i - rv_j}{|rv_i - rv_j|} \tag{6}$$

According to the computed desired location $rv'_m$ for $rv_m$, the force $f_i$ for $rv_m$ should be $\overrightarrow{rv_m rv'_m}$. Then displacement of each node in the linear network is computed using the snake model according to their accepted forces. However, running the algorithm only once cannot guarantee that all circles are in suitable locations (Liu et al., 2014). Therefore, the region centroids are iteratively displaced according to the forces until convergence, in which boundary nodes are also displaced along the displacement of region centroids by using the Snake model. Considering that some region centroids may accept large forces, and using the Snake model can only guarantee the stability of the solution results when the forces are small. Therefore, we limit the maximum force on the $\{rv_m\}_{m=1}^{M}$ during each displacement process. Suppose the force for $rv_m$ is $f_i$ according to Eq.(5), the maximum force for a region centroids in $\{rv_m\}_{m=1}^{M}$ is $|f_{\max}|$, the threshold for all forces are defined as $T_f$. If $|f_{\max}| > T_f$, then we adjust $f_i$ for $rv_m$ according to Eq. (7).

$$f_i' = f_i \times \frac{T_f}{|f_{\max}|} \tag{7}$$

### 4.4 Step 3. Using the snake model

We apply the Snake model to displace the region centroids $\{rv_m\}_{m=1}^{M}$ and boundary node $\{bv_n\}_{n=1}^{N}$ simultaneously. But applying the Snake model once cannot guarantee a satisfactory result and an iterative process needs to be performed. The stop conditions for the Snake model. are defined as follows.

**Condition 1**: The iterative process stops if the region centroids $\{rv_m\}_{m=1}^{M}$ are equidistant (grid-like). Satisfying the situation means the nodes don't need to be displaced and can be ruled by setting a threshold for the maximum displacement ($\max(d_i)_t$) of each node at step $t$. If $\max(d_i)_t \le \varepsilon$, then the iterative process stops; $\varepsilon$ is a small constant.

**Condition 2**: The iterative process stops if the region centroids $\{rv_m\}_{m=1}^{M}$ cannot be more equidistant (grid-like). This can be ruled based on the maximum displacement of each point during two adjacent steps as $\max(d_i)_{t-1}$ and $\max(d_i)_t$. If $\max(d_i)_t > \max(d_i)_{t-1}$, then the iterative process stops.

**Condition 3**: The iterative process stops if a maximum step ($T_s$) is reached and can be set by users.

The iterative process is implemented as Algorithm 1.

**Algorithm 1. Iterative process for hierarchical optimization**

**Input**: the regions are $\{reg_m\}_{m=1}^{M}$, the boundary polygon of $\{reg_m\}_{m=1}^{M}$ is $P_{\text{regs}}$, the maximum iterative step is $T_s$

**Initialize**: the iterative step as $IS \leftarrow 0$, the maximum displace displacement as $\max(d_i) \leftarrow 0$;

Compute the centroids of $\{reg_m\}_{m=1}^{M}$ as $\{rv_m\}_{m=1}^{M}$, get the boundary nodes of $P_{\text{regs}}$ as $\{bv_n\}_{n=1}^{N}$, and construct the proximity graph as $G$

**Do**

Force computation for $\{rv_m\}_{m=1}^{M}$ in $G$

Compute the displacement vector for each node in $G$ with the Snake model, the maximum force is $\max(d_i)_t$;

**If** $\max(d_i)_t < \max(d_i)$ **Then End loop**

**Else**

Update the coordinates $\{bv_n\}_{n=1}^{N}$ and $\{rv_m\}_{m=1}^{M}$ based on the displacement vector, update $G$;

$IS \leftarrow IS + 1$;

**While** ( $\max(d_i)_t > \varepsilon$ AND $IS \le T_s$ )

**Return** $P_{\text{regs}}$ and $\{rv_m\}_{m=1}^{M}$

### 4.5 Step 4. Grid adaptation and assignment

Given the region centroids as $\{rv_m\}_{m=1}^{M}$ and an associated boundary polygon as $P_{\text{regs}}$, the bounding box of $P_{\text{regs}}$ is $P_{\text{box}}$ and denoted as [$X_{\max}$, $X_{\min}$, $Y_{\min}$, $Y_{\max}$]. We wish to find a set of $M$ grids that fit $P$ as much as possible. To achieve this, we lay grids with step size as defined in Eq. (5) over $P$, the star origin is defined as ($X_{\min}$, $Y_{\max}$). Suppose the grids that intersect $P$ as $\{G_i\}_{i=1}^{M'}$, where $M' \ge M$. We need to get $M$ grids from $M'$ grids. We assign a cost ($C_{\text{grid}}$) for each grid in $\{G_i\}_{i=1}^{M'}$ as Eq. (8).

$$C_{\text{grid}} = \begin{cases} A_{overlap} / A & (\text{if the grid centroid is inside } P_{regs}) \\ A_{overlap} / A + 1 & (else) \end{cases} \tag{8}$$

We then sort the grids in $\{G_i\}_{i=1}^{M'}$ in descending order by cost, and select the top $M$ grids.

Then we assign each grid to a region. The region centroids after displacement denote as $\{rv_m\}_{m=1}^{M}$, and we get $M$ grids after the grid adaptation. We seek a 1-to-1 assignment that minimizes changes in centroid position according to McNeil and Hale (2017). Specifically, we seek a permutation $\phi$ of $\{R_i\}_{i=1}^{M}$ that minimizes the cost function:

$$f(\phi) = \sum_{i=1}^{M} \left\| rv^{new}{}_i - t_{\phi(i)} \right\|^2 \tag{9}$$

where $t_i$ is the centroid of $R_i$. We solve the assignment problem (i.e. find the optimal $f$) using the Hungarian method (the Kuhn-Munkres algorithm).

### 4.6 The strategies to create multiple candidates

Although physical geography plays a crucial role in determining the perceptual significance of individual geographic features and their relationships, task-specific factors and the end users' cultural and demographic knowledge of the area also contribute to the importance of preserving specific features of the original map. Therefore, users may have different preferences for the grid maps. To address this issue, we also propose two strategies according to McNeill and Hale (2017) that offer users options to create multiple candidate grid maps based on the criteria outlined above. This allows users to select an appropriate map for their specific task.

(1) **Strategy 1**. Add Gaussian noise to the region centroids

The region centroids may have different densities, a standard deviation of the Gaussian noise can be specified as a proportion of the mean distance of a centroid from its neighbors according to McNeil and Hale (2017). Adding unique noise to the original centroids allows us to create as many candidate sets of noisy centroids as desired.

(2) **Strategy 2**. Start with another origin in grid adaptation

The grid adaptation approach discussed in Section 4.5 provides a means of generating the grid arrangement. By shifting the grid's origin (either horizontally or vertically) before fitting the grid points to the boundary, we can create multiple sets of grids that can be presented as options for users to choose from.

# 5 Experiment

## 5.1 Implement details

### 5.1.1 *Dataset*

Two datasets were applied to validate the proposed approach.

**Dataset A**. This dataset includes 49 states of the United States of America, excluding Alaska and Hawaii (data source: https://hub.arcgis.com/), which is also a benchmark dataset for grid map production that has been previously used in the literature (McNeill and Hale, 2017).

**Dataset B**. This dataset includes 75 cities in Central China (data source: http://bzdt.ch.mnr.gov.cn/).

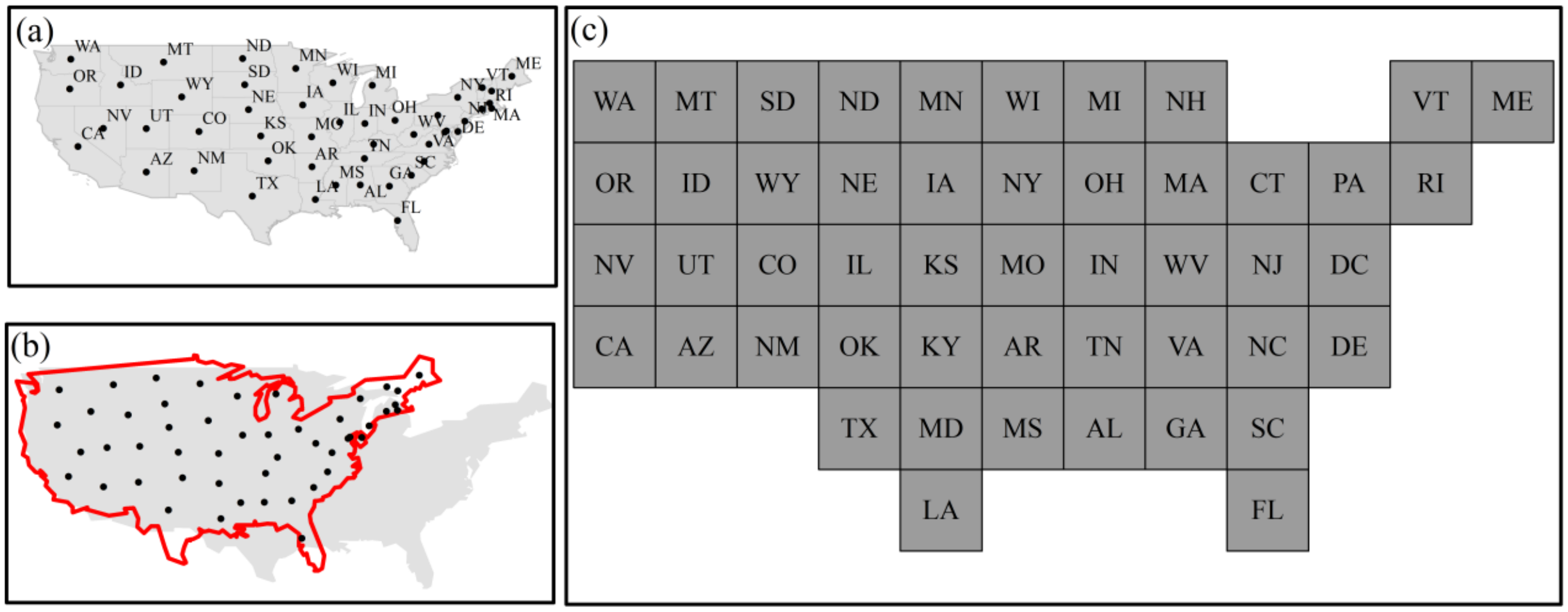

Figure 3. The grid map generated by using the proposed approach for Dataset A.

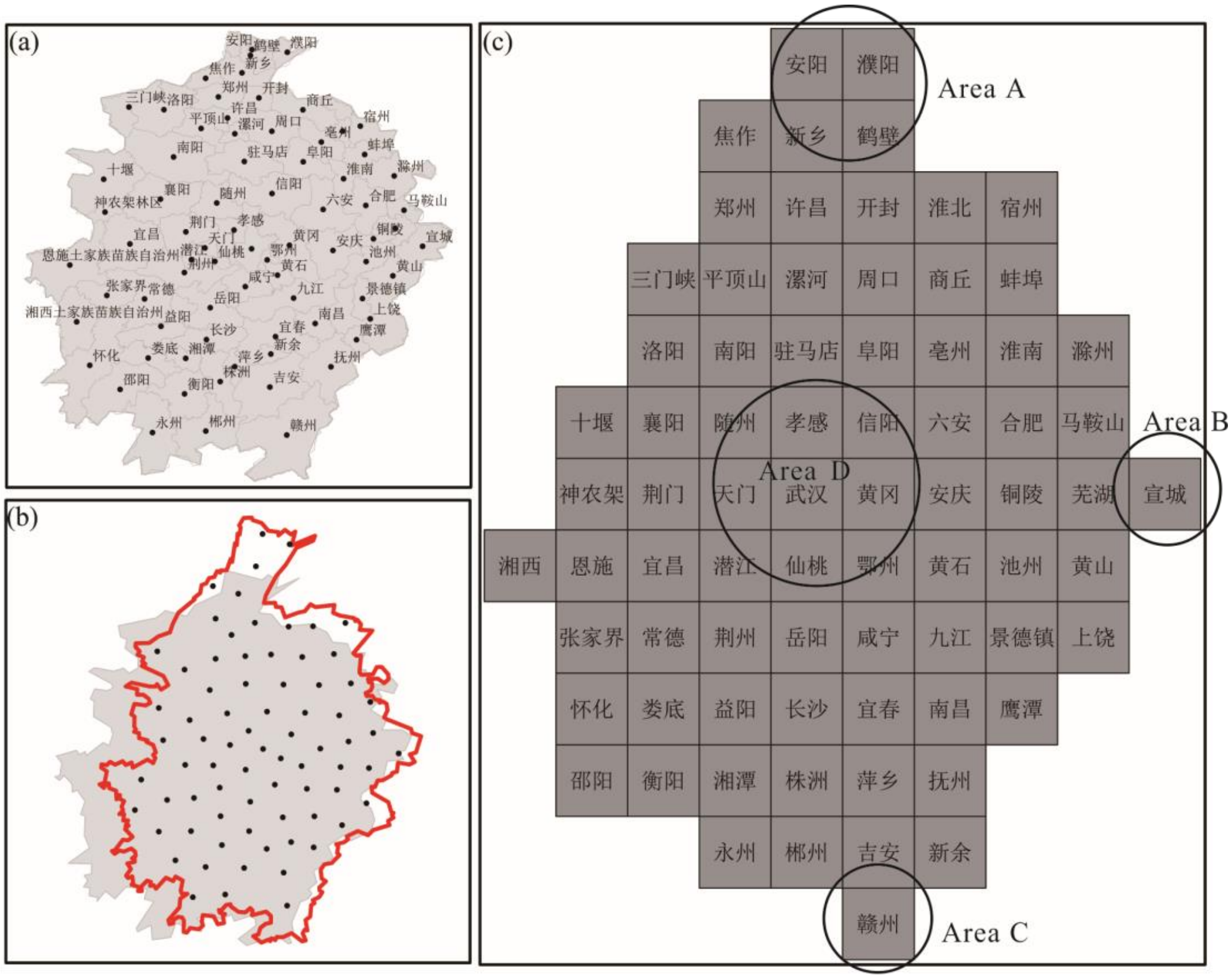

Figure 4. The grid map generated by using the proposed approach for Dataset B.

5.1.2 *Evaluation metric*

The resulting quality was evaluated according to the quality criteria in Section 3. The evaluation metrics were employed as delineated below.

(1) Local location cost ($C_{location}$): This metric was used to evaluate the local location maintenance criteria and was determined by calculating the mean distance between the transformed region centroids and the centroids of the grids to which they were assigned (McNeil and Hale, 2017).

(2) Adjacency cost ($C_{adjacent}$): This metric was used to evaluate the topology relation maintenance criteria and was quantified by the ratio of preserved adjacency relations (McNeil and Hale, 2017).

(3) Relative orientation cost ($C_{orientation}$): This metric was used to evaluate the relative orientation maintenance criteria and was quantified by the average values of the difference in direction angles for the links between the centers of neighboring grids to their original corresponding regions (Wei et al., 2023). Specifically, these links were the edges in the constructed linear network.

(4) Shape cost ($C_{shape}$): This metric was used to evaluate the global shape maintenance criteria and was determined by a shape similarity index, which compared the grid map boundary to its original boundary using Fourier transform methodology (Wei et al., 2021).

5.1.3 *Experimental environment and parameter setting*

We implemented our proposed approach using C# code in the ArcEngine 10.2 environment. The experiments were conducted on a personal computer equipped with an AMD Ryzen 7-7840HS Radeon 780M Graphics @3.80 GHz CPU and 16GB RAM.

In configuring our approach, three crucial parameters require specification. (1) The maximum force threshold($T_f$) in force computation, set to $T_f$=0.5cm (graphic distance) based on positional accuracy constraint for map objects (Liu et al., 2014). (2) The parameters $\alpha$ and $\beta$ that govern the elasticity and stiffness of the snake model, and Liu et al. (2014) have extensively analyzed the effects of these parameters. Following their conclusions, we set $\alpha = \beta = 100000$ to ensure consistent elasticity and stiffness properties along the line. Additionally, these parameters are updated based on the maximum force in each iteration, with the updating strategy outlined in Liu et al. (2014). (3) The Snake algorithm iteration parameter($T_s$), set to $T_s$=30 according to McNeil and Hale (2017).

Additionally, beyond the default origin setting outlined in Section 4.5, we applied **Strategy 2** from Section 4.6 to create four layouts. This involved shifting the origin horizontally or vertically by 0.5 grid size (*S*) in separate instances. The most suitable layout, determined through a Topsis approach (Çelikbilek et al., 2020) considering the disparate dimensions of the four evaluation metrics, was selected as our final result. Considering the need for expression details and computational efficiency, geographical boundaries were simplified using the POINT_REMOVE method under the Simplify Polygon tool in ArcGIS 10.2. The simplification threshold was set to 0.1 degrees, and the simplified boundaries are depicted in Figures 3(b) and 4(b).

### 5.2 Evaluation and comparison

5.2.1 *The existing approach for comparison*

Grid maps for comparisons were generated from the same two datasets using the established approach outlined by McNeill and Hale (2017). To diversify the comparison, we also implemented **Strategy 2** from Section 4.6, generating four layouts by shifting the origin horizontally or vertically with a 0.5 grid size (*S*) independently. The optimal layout, chosen based on the evaluation criteria, was selected for subsequent comparisons.

5.2.2 *The evaluation and comparison*

The results of our approach are presented in Figures 3 and 4, and a detailed statistical analysis of the proposed and existing approaches is provided in Table 1. Further insights into the efficiency of the proposed algorithm will be discussed in Section 6 Discussion.

5.2.2.1 Quantitative analysis

Analyzing Table 1 yields the following observations. For Dataset A, $C_{adjacent}$ and $C_{shape}$ in the proposed approach compared to the existing approach show an increase of 0.01 and 0.02, respectively, $C_{orientation}$ records a reduction of 1.58°, while $C_{location}$ witnesses an increase by 6.12 *$10^6$m. For Dataset B, $C_{orientation}$ in the proposed approach compared to the existing approach show a reduction of 6.48°, $C_{shape}$ records an increase of 0.03, while $C_{location}$ witnesses a reduction by 71.27 *$10^6$m and $C_{adjacent}$ records an increase of 0.12.

These observations suggest that the proposed approach excels in maintaining relative orientation and global shape. However, it might introduce higher variations in local location between grids and their corresponding region centroids. This tendency could be attributed to our approach's simultaneous transformation of nodes and boundaries, where the relations between regions are modeled as a graph to govern the global structure. Regarding adjacent relation maintenance, neither approach exhibits a distinct advantage.

Table 1. Statistical results on the effectiveness of the proposed approach and the existing approach.

| | Proposed approach | | Existing approach | |
|---|---|---|---|---|
| | Dataset A | Dataset B | Dataset A | Dataset B |
| $C_{location}\uparrow$ | 38.01 | 224.10 | 31.89 | 152.83 |
| $C_{adjacent}\uparrow$ | 0.83 | 0.63 | 0.82 | 0.75 |
| $C_{orientation}\downarrow$ | 27.15 | 24.15 | 28.73 | 30.63 |
| $C_{shape}\uparrow$ | 0.82 | 0.84 | 0.80 | 0.81 |

5.2.2.2 Qualitative analysis

A qualitative analysis was conducted on the grid map for Dataset B generated by the existing approach, and the visualization result is presented in Figure 5. A comparison between Figures 4 and 5 reveals the following observations.

As for the transformed nodes, the proposed approach achieves a more equidistant (grid-like) node layout, while the existing approach may generate dense areas, exemplified by Area B in Figure 5. The disparity arises from the fact that the existing approach only considers the adjacent relation between regions for node transformation, whereas our approach simultaneously considers adjacent relations and relative orientation relations between regions.

As for the transformed boundary, the boundary produced by the proposed approach exhibits no self-intersections, and no nodes extend beyond the boundary. In contrast, the existing approach results in three self-intersections (Area A in Figure 5(b)), and five nodes lie outside the boundary (Areas C and D in Figure 5(b)). The reason is that our approach's simultaneous transformation of nodes and boundaries, with relations between regions modeled as a linear network, contributes to this improved outcome. In contrast, the existing approach transforms nodes and boundaries separately.

As for the produced grid map, by zooming into specific local regions A, B, C, and D in Figure 4, it becomes evident that our proposed approach excels in preserving relative orientation and overall shape. For instance, in Area A, our method successfully maintains the relative positioning of the four cities (安阳, 濮阳, 新乡, and 鹤壁) similar to the original map. In contrast, the traditional method introduces an unrelated city (宿州) to replace 濮阳 due to the self-intersection of the transformed boundary, which is geographically distant from the other cities. Moving to Areas B and C, it can be observed that 宣城 and 赣州 occupy peripheral positions on the boundary, accurately reflecting the global shape. Our approach ensures that these cities remain at the border extremities in the grid map. Conversely, in the grid map produced by the existing method, these cities are misplaced, failing to adhere to the original boundary shape. Focusing on the central hub of 武汉 in Area D, a crucial city in central China, it is noteworthy that our grid map preserves the adjacency relationships with neighboring cities such as 鄂州, 黄冈, 咸宁. Both maps maintain these adjacency relationships, further validating the effectiveness of our proposed approach.

The qualitative analysis indicates that our approach achieves superior transformation of nodes and boundaries, resulting in a grid map that better preserves relative orientation and global shape.

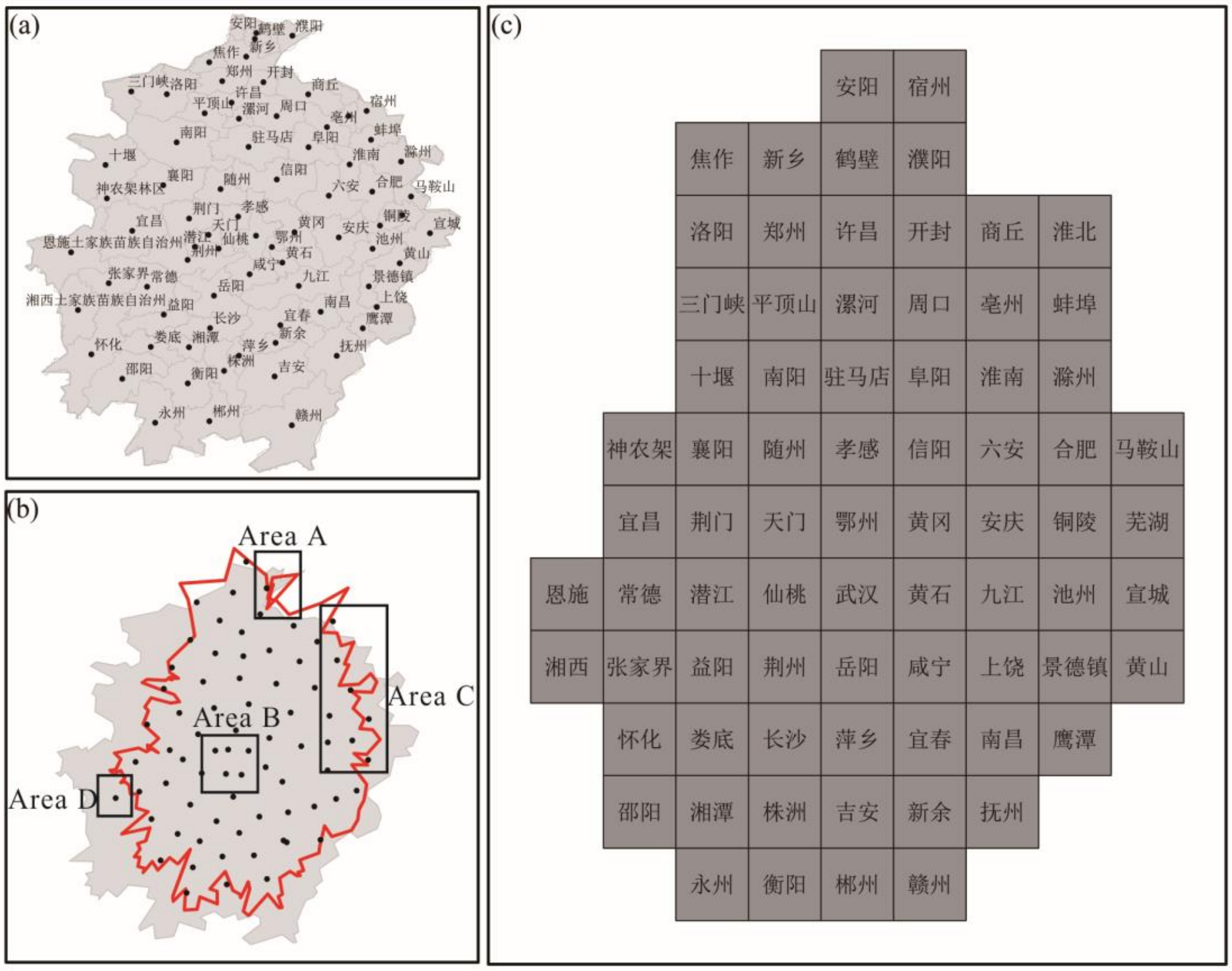

Figure 5. The grid map generated by using the approach according to McNeil and Hale (2017) for Dataset B.

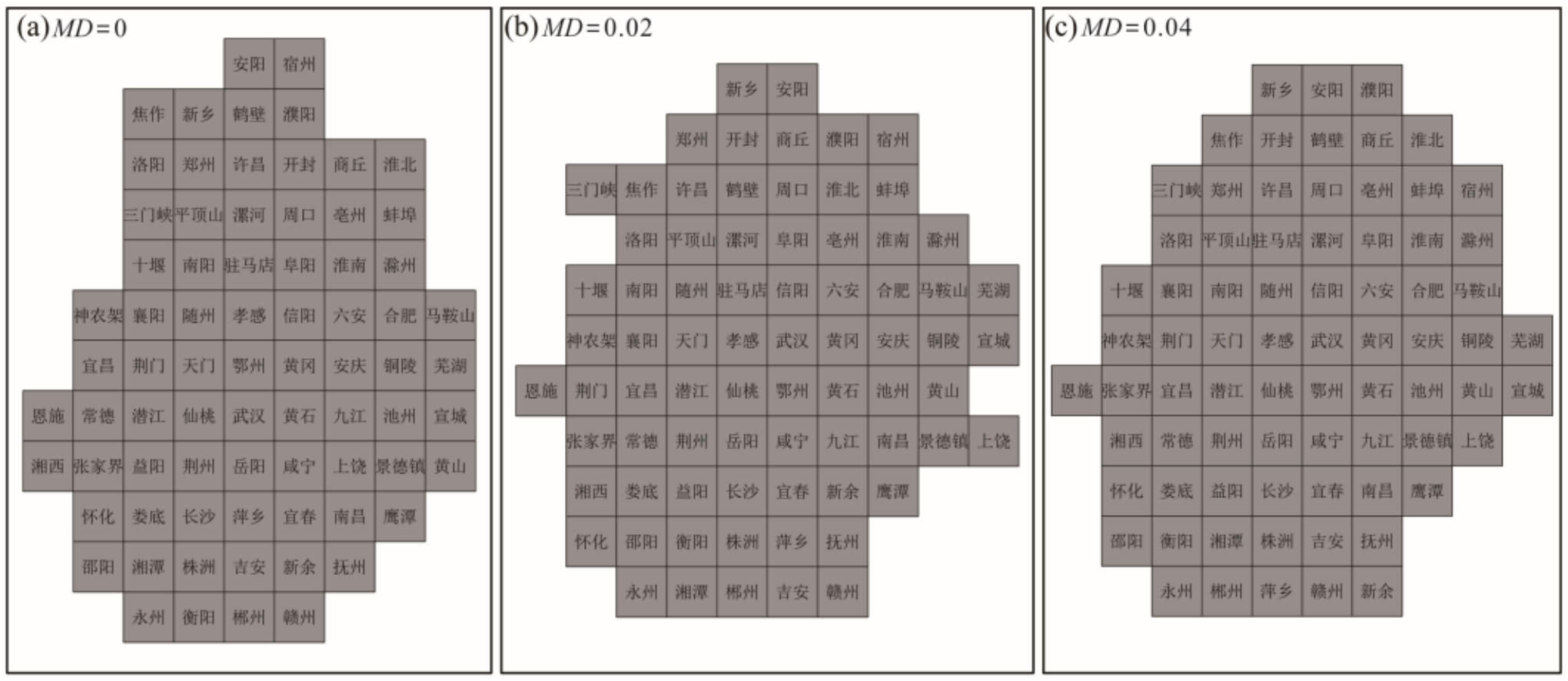

Figure 6. The grid maps generated by using the proposed approach for Dataset B with different Gaussian noises.

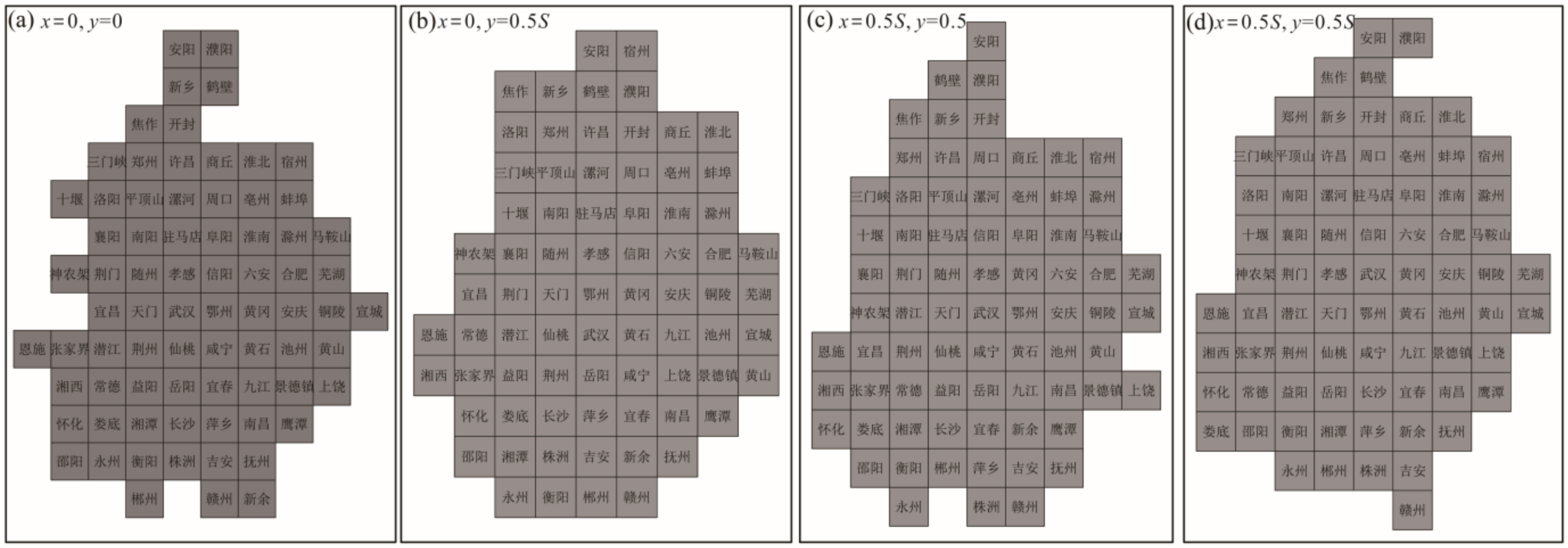

Figure 7. The grid maps generated by using the proposed approach for Dataset B with different origin shifts.

## 6 DISCUSSION

### 6.1 Impact of the degree of simplification of geographical boundaries on experimental results and efficiency analysis

The time consumption ($t$) will increase with the augmentation of nodes in the built linear network due to the time-consuming matrix operations within the Snake algorithm. Geographical boundaries, dictated by expressive requirements, exhibit varying levels of detail, resulting in different node counts. In our approach, we simplified geographical boundaries using the POINT_REMOVE method under the Simplify Polygon tool in ArcGIS 10.2 with a simplification threshold of 0.1 degrees. Therefore, we also employed the POINT_REMOVE simplification method to generate distinct levels of detail for geographical boundaries by applying different threshold values of 0.5 degrees, 0.2 degrees, 0.1 degrees, 0.05 degrees, and 0.02 degrees. Subsequently, the method proposed in this paper is implemented to generate grid maps based on these geographical boundaries to explore its efficiency concerning experimental results.

In this experiment, the number of region centroids is held constant at 75, and as the geographical boundaries become more detailed, the count of boundary nodes increases. We applied region centroids count ($N_{region}$), boundary node count ($N_{boundary}$), the total node count ($N_{sum}$), and the ratio of centroids count to boundary node count ($N_{ratio}$) to quantify the level of geographical boundary simplification. The time consumption ($t$) was applied to evaluate the algorithm's efficiency. The results are summarized in Table 2.

As shown in Table 2, when the total node count approaches 200, the time consumption is nearly 1 minute. However, as the total node count surpasses 400, the algorithm's time consumption exceeds 10 minutes. This escalation is attributed to the Snake model, which involves extensive matrix calculations, and its computational complexity experiences exponential growth with an increase in the number of points. Consequently, as the expression of geographical boundaries becomes more intricate, the algorithm's efficiency decreases. For practical applications, considering algorithm efficiency, it is recommended to maintain the ratio of region centroids to boundary nodes ($N_{ratio}$) at approximately 0.5, with a total node count nearing 200. This provides valuable guidance for determining the requisite level of detail for expressing geographical boundaries.

Table 2. Statistical results on the efficiency of the proposed approach with boundaries in different levels of detail.

| Simplification parameter | 0.5 | 0.2 | 0.1 | 0.05 | 0.02 |
|---|---|---|---|---|---|
| $N_{region}$ | 75 | 75 | 75 | 75 | 75 |
| $N_{boundary}$ | 28 | 72 | 136 | 291 | 851 |
| $N_{sum}$ | 103 | 147 | 211 | 366 | 926 |
| $N_{ratio}$ | 2.68 | 1.04 | 0.55 | 0.26 | 0.09 |
| $t$ | 8.94 | 23.54 | 66.40 | 660.13 | 5784.07 |

### 6.2 The strategies analysis for creating multiple candidates

We implemented two strategies in our approach to generate multiple grid map candidates, and this section scrutinizes the impact of these strategies on the experimental outcomes.

#### 6.2.1 Analysis of strategy 1

Strategy 1 introduces Gaussian noise to the region centroids, where even minute centroid perturbations can markedly influence the resulting shape (as illustrated in Figure 6). Gaussian noise was applied with mean deviation ($MD$) set as 0, 0.02, and 0.04 to investigate the impact on experimental results. Results are detailed in Figure 6 and Table 3. Figure 6 demonstrates the successful production of three distinct grid maps through the application of varied Gaussian noise. Table 3 reveals that the grid map with $MD$=0 attains the minimum value in $C_{location}$(38.01*$10^6$m) and the maximum value in $C_{adjacent}$(0.83), whereas the grid map with $MD$=0.04 achieves the minimum value in $C_{orientation}$(26.41°) and the maximum value in $C_{shape}$ (0.92).

#### 6.2.2 Analysis of strategy 2

Strategy 2 involves shifting the grid's origin, either horizontally or vertically, before aligning the grid points with the boundary. Origin shifts were conducted horizontally or vertically with 0.5 grid size ($S$), generating four grid maps for evaluation. Results are presented in Figure 7 and Table 3. Figure 7 illustrates the successful generation of four distinct grid maps with different origin shifts. Table 3 shows that the grid map with $x$=0, $y$=0.5$S$ attains the minimum value in $C_{location}$(38.01*$10^6$m) and the maximum value in $C_{adjacent}$(0.83), while the grid map with $x$=0.5$S$, $y$=0.5$S$ achieves the minimum value in $C_{orientation}$(25.85°) and the grid map with $x$=0, $y$=0.5$S$ attains the maximum value in $C_{shape}$(0.92).

These observations for both strategies underscore that strategies with varying parameters can effectively generate satisfactory grid maps, each potentially excelling in specific aspects. The selection of an appropriate strategy can thus be tailored to user preferences and demands in practical applications.

Table 3. Statistical results on the effectiveness of the proposed approach with different strategies.

| | $C_{location}$↑ | $C_{adjacent}$↑ | $C_{orientation}$↓ | $C_{shape}$↑ |
|---|---|---|---|---|
| $x$=0, $y$=0, $MD$=0 | 43.93 | 0.79 | 25.95 | 0.80 |
| $x$=0, $y$=0.5$S$, $MD$=0 | 38.01 | 0.83 | 27.15 | 0.82 |
| $x$=0.5$S$, $y$=0, $MD$=0 | 45.33 | 0.77 | 29.15 | 0.84 |
| $x$=0.5$S$, $y$=0.5$S$, $MD$=0 | 42.74 | 0.79 | 25.85 | 0.82 |
| $x$=0, $y$=0.5$S$, $MD$=0.02 | 44.48 | 0.74 | 27.36 | 0.86 |
| $x$=0, $y$=0.5$S$, $MD$=0.04 | 40.33 | 0.79 | 26.41 | 0.92 |

### 6.3 Parameter analysis

Our approach involves configuring two parameters, one of which governs the iterative process of the Snake algorithm and can be adjusted interactively by users. In this analysis, we specifically focus on the influence of the other parameter, the force threshold ($T_f$), which regulates the maximum force in force computation. For our baseline setting (Figure 3), we selected $T_f$=0.5cm (graphical distance) in Section 5.1.2. We conducted a comparative evaluation by not setting a threshold for forces in force computation. The outcomes related to transformed nodes and boundaries are presented in Figure 8.

Figure 8 underscores the challenges encountered when forces in force computation are unrestricted, resulting in suboptimal transformed nodes and boundaries. Specifically, when the iterative step (*IS*) reaches 25, the region centroids fail to achieve a more equidistant (grid-like) layout, as defined in Section 4.4. However, the transformed nodes extend beyond the transformed boundary, exemplified in Area A in Figure 5. When *IS*=30, the consequence is a self-intersected transformed boundary. This occurrence is attributed to the requirement for relatively small displacements to maintain result stability when employing the Snake displacement method for problem-solving (Wang et al., 2018). These findings underscore the essential nature of setting a threshold for force ($T_f$) to ensure the stability of results.

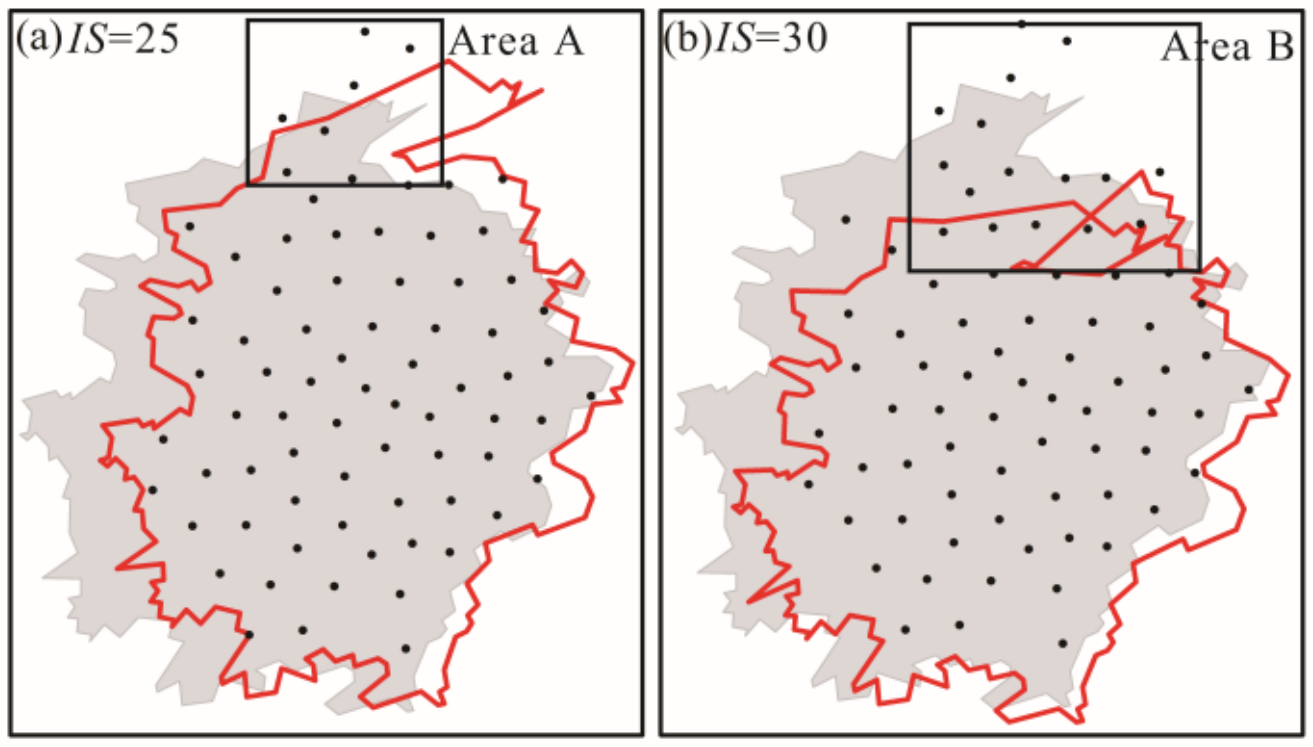

Figure 8. The transformed region centroids and boundaries without maximum force threshold.

### 6.4 Limitation analysis

While the proposed approach yields satisfactory results under the defined quality constraints outlined in Section 3, several limitations also need to be addressed and improved in future studies.

(1) The current approach primarily leverages size and shape as graphic variables, with the potential incorporation of color as an additional variable remaining unexplored. Future research endeavors should investigate the role of color, enabling it to represent regional properties and offering a more nuanced perspective. For instance, assigning different colors to various regions for quantity visualization could enhance the distinctiveness of the visual representation.

(2) Considering the intricacies often encountered in real-world boundaries, such as the existence of voids or isolated regions (Yang et al., 2019), our approach may necessitate further development or the adoption of novel strategies to address these scenarios adeptly. One prospective method could entail partitioning the shape into multiple segments for subsequent processing, as recommended by Meulemans et al. (2020). Subsequently, our proposed approach could be applied to each polygon individually to produce a sub-grid map.

(3) As discussed in Section 6.1, when the total node count for grid map production exceeds 400, the algorithm's time consumption exceeds 10 minutes. To enhance scalability with larger datasets, efficiency improvements are needed. The efficiency of our proposed approach is mainly constrained by the Snake model, which involves extensive matrix calculations. Its computational complexity experiences exponential growth with an increase in the number of points. This challenge can be addressed by dividing large datasets into smaller parts or employing more efficient matrix computation strategies.

(4) The provided code in our GitHub repository is developed using C#. However, many commercial or open-source software products, such as D3 or Vega (Bostock et al., 2011; Satyanarayan et al., 2016), are primarily developed for JavaScript, offering better usability and accessibility. Therefore, to support a wider application of our approach, efforts should be made to provide a more user-friendly implementation.

## 7 Conclusion

To generate grid maps for geospatial data visualization, we conceptualize the construction process as a displacement problem within map generalization. Subsequently, we employ the Snake displacement algorithm, initially designed for cartographic generalization, to iteratively transform nodes and boundaries simultaneously, resulting in the creation of grid maps. During the application of the Snake algorithm, we conceptualize the relations between regions as a linear network, strategically controlling its structure in the subsequent grid map production. Experimental findings demonstrate that, in comparison to existing methods, the proposed approach excels in preserving the relative orientation between regions and overall shape. However, it may exhibit an increase in local location deviations. Additionally, two strategies, aligned with McNeill and Hale's (2017) recommendations, are presented to cater to users' varied preferences in generating diverse grid maps. Notwithstanding its effectiveness, it's noteworthy that the Snake algorithm involves extensive matrix calculations, leading to a decline in computational efficiency with larger datasets. Future works will focus on two key aspects: (1) Incorporating more graphic variables, such as colors, to enhance the informativeness of visualizations; and (2) Enhancing algorithm efficiency to ensure scalability with larger datasets.